\documentclass[10pt,twocolumn,letterpaper]{article}

\usepackage[pagenumbers]{cvpr} %

\usepackage{multirow}
\usepackage{amsmath}
\usepackage[dvipsnames,table]{xcolor}
\usepackage{graphics}
\usepackage{graphicx}
\usepackage{amssymb}
\usepackage{booktabs}
\usepackage{makecell}
\usepackage{array}
\usepackage{wrapfig}
\usepackage{multirow}
\usepackage{bbm}
\usepackage{float}
\usepackage{graphicx}
\usepackage{perpage}
\usepackage{enumitem}
\usepackage{algpseudocode}
\usepackage[ruled,vlined]{algorithm2e}
\usepackage{extarrows}
\usepackage{amssymb}
\usepackage{booktabs}
\usepackage{marvosym}
\usepackage{pifont}
\usepackage{gensymb}
\usepackage{adjustbox}
\usepackage{ragged2e}
\usepackage{siunitx}
\usepackage{ifsym}

\newcommand{\ourmethod}{DCNv4}
\newcommand{\ournet}{FlashInternImage}

\newcommand\blfootnote[1]{%
    \begingroup
    \renewcommand\thefootnote{}\footnote{#1}%
    \addtocounter{footnote}{-1}%
    \endgroup
}

\definecolor{cvprblue}{rgb}{0.21,0.49,0.74}
\usepackage[pagebackref,breaklinks,colorlinks,citecolor=cvprblue]{hyperref}
\usepackage{microtype}

\title{Efficient Deformable ConvNets: Rethinking Dynamic and Sparse
    \\ Operator for Vision Applications}

\author{
    Yuwen Xiong$^{*1,2}$ \quad Zhiqi Li$^{*3,2}$ \quad Yuntao Chen$^{*4}$ \quad Feng Wang$^{*5}$ \\
    Xizhou Zhu$^{5,6}$ \quad Jiapeng Luo$^{6}$ \quad Wenhai Wang$^{7,2}$ \quad Tong Lu$^3$ \quad Hongsheng Li$^7$\\
    Yu Qiao$^2$ \quad Lewei Lu$^6$ \quad Jie Zhou$^5$ \quad Jifeng Dai$^{5, 2}\textsuperscript{\Letter}$\\\\
    $^1$University of Toronto \quad\quad\quad  $^2$OpenGVLab, Shanghai AI Laboratory \\
    $^3$Nanjing University \quad\quad\quad $^4$CAIR, HKISI, CAS  \quad\quad\quad $^5$Tsinghua University \\
    $^6$SenseTime Research\quad\quad\quad $^7$The Chinese University of Hong Kong\\
    \\\url{https://github.com/OpenGVLab/DCNv4}
}

\begin{document}
\maketitle
\blfootnote{
    * Equal contribution\\
    \indent\indent\Letter~Corresponding author (daijifeng@tsinghua.edu.cn)
}
\begin{abstract}

We introduce Deformable Convolution v4 (DCNv4), a highly efficient and effective operator designed for a broad spectrum of vision applications. DCNv4 addresses the limitations of its predecessor, DCNv3, with two key enhancements: 1. removing softmax normalization in spatial aggregation to enhance its dynamic property and expressive power and 2. optimizing memory access to minimize redundant operations for speedup. These improvements result in a significantly faster convergence compared to DCNv3 and a substantial increase in processing speed, with DCNv4 achieving more than three times the forward speed.
DCNv4 demonstrates exceptional performance across various tasks, including image classification, instance and semantic segmentation, and notably, image generation. 
When integrated into generative models like U-Net in the latent diffusion model, DCNv4 outperforms its baseline, underscoring its possibility to enhance generative models.
In practical applications, replacing DCNv3 with DCNv4 in the InternImage model to create FlashInternImage results in up to 80\% speed increase and further performance improvement without further modifications.
The advancements in speed and efficiency of DCNv4, combined with its robust performance across diverse vision tasks, show its potential as a foundational building block for future vision models.

\end{abstract}
\vspace{-0.2cm}
\section{Introduction}
\label{sec:intro}

In the field of computer vision, there is an ongoing debate about whether convolutional networks (ConvNets) or Transformers offer superior performance. 
In recent years, Transformer models~\cite{dosovitskiy2020image,liu2021swin,zhai2022scaling} have achieved remarkable results in large vision models with the attention mechanism, showing the potential to overtake ConvNets.
However, recent works such as InternImage~\cite{wang2023internimage} and ConvNeXt~\cite{liu2022convnet} demonstrate that ConvNet-based vision models retain robust performance, efficiency, simplicity, and suitable inductive bias for various downstream tasks~\cite{he2017mask, xiao2018unified}. 
Notably, in domains like image generation~\cite{rombach2022high, saharia2022photorealistic}, convolution remains the preferred approach. 
This situation brings to light the enduring value of convolution-based approaches.

\begin{figure}[t]
    \centering
    \subfloat[]{
\includegraphics[width=0.59\linewidth]{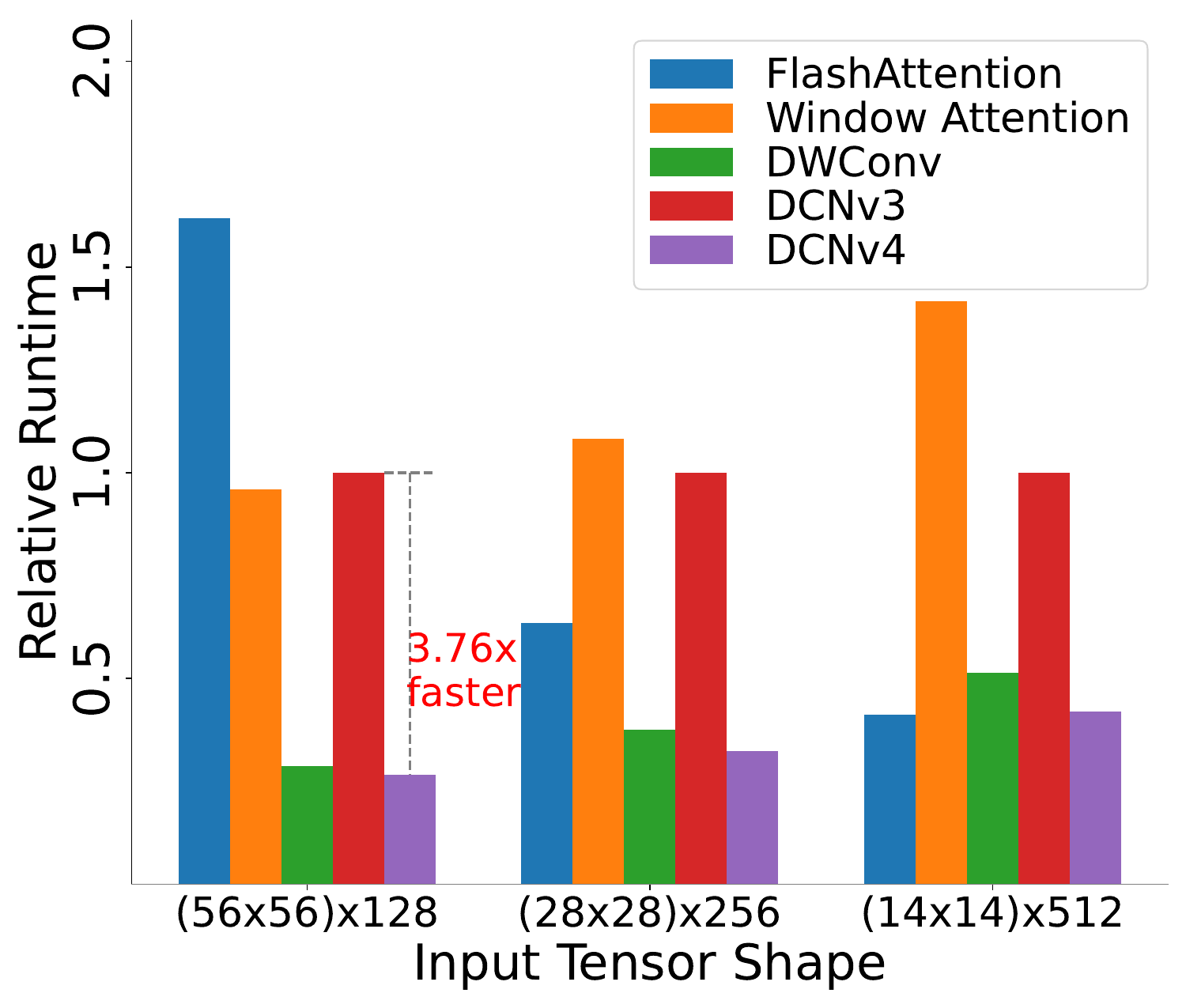}
\label{fig:teaser_1}
}
     \subfloat[]{\includegraphics[width=0.39\linewidth]{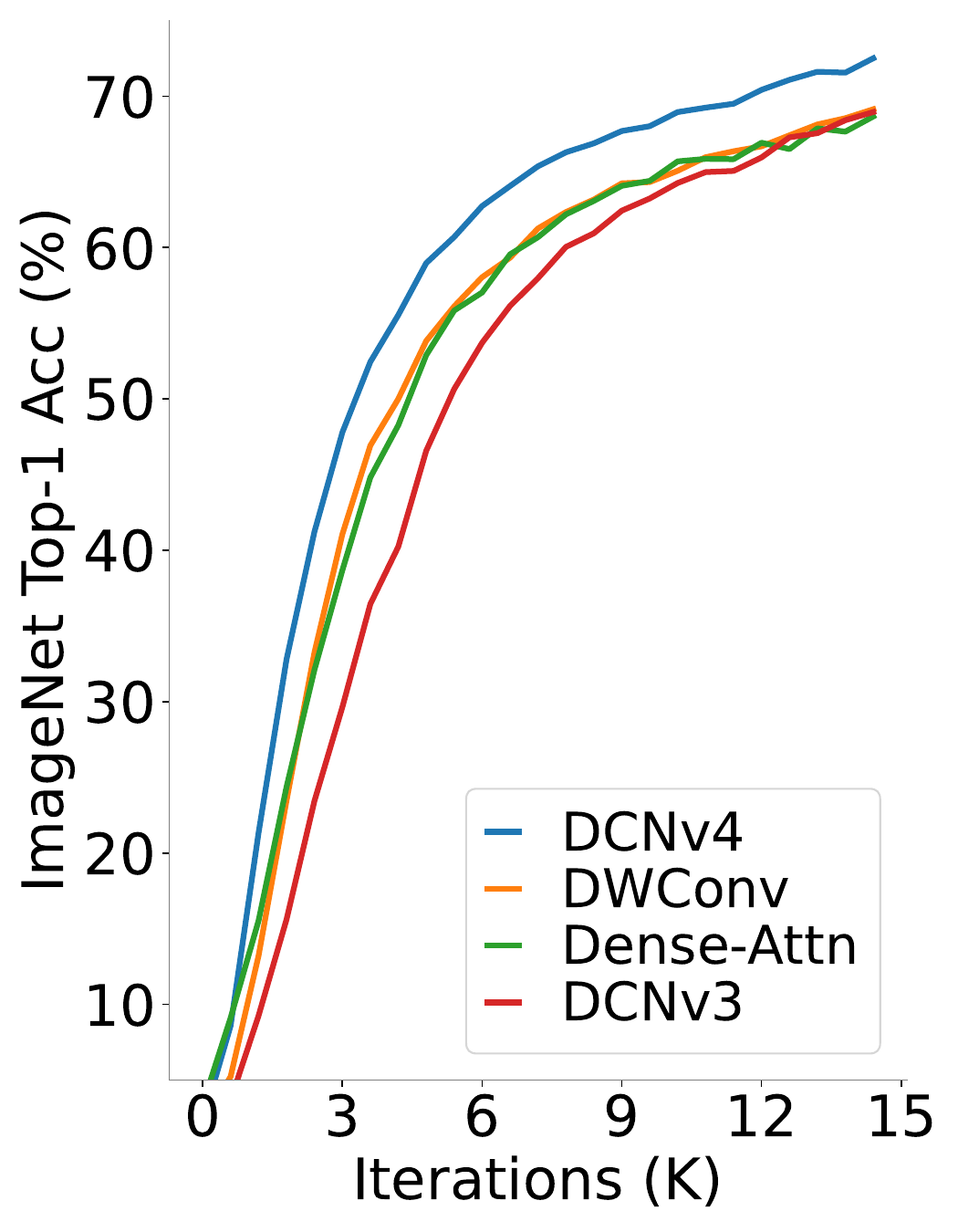}
    \label{fig:teaser_2}}
    \vspace{-0.3cm}
    \caption{(a) We show relative runtime with DCNv3 as the baseline. \ourmethod{} shows {\bf significant speedup} over DCNv3, and surpasses other common vision operators. (b) With the same network architecture, DCNv4 {\bf converges faster} than other operators, while DCNv3 falls behind in the initial training phase.}
    \vspace{-0.6cm}
\end{figure}

Building on convolution's strengths, Deformable Convolution v3 (DCNv3) – the core operator of the advanced ConvNet model InternImage – innovatively combines a sparse attention mechanism with convolution: it processes each output location in a sliding window manner with a small window size (\eg $3\times 3=9$ points) which acts as a local, sparse operator like convolution, while dynamically samples point with an adaptive range and aggregates the spatial features with input-dependent attention weights.
With its small window size and ConvNet inductive bias, DCNv3 is expected to achieve a faster convergence rate and lower inference latency, especially when compared to dense global~\cite{dosovitskiy2020image} or local window-based~\cite{liu2021swin} attention methods.

Despite these advantages, DCN has not become the go-to solution for vision backbone models.
This observation led us to investigate the lingering limitations of the DCN operator.
The first thing we notice is the running speed. 
The slow speed of DCN is known to be a long-standing problem~\cite{ahn2020efficient}, as it introduces extra overhead on sampling non-nearby locations, making it not fit modern convolution algorithms. 
Our comparative analysis, illustrated in Fig.~\ref{fig:teaser_1}, reveals that DCNv3 can be slower than a properly optimized dense global attention~\cite{dao2022flashattention}, highlighting the need for further optimization.
Moreover, we find DCNv3 even converges slower than global attention at the initial backbone training phase, as shown in Fig.~\ref{fig:teaser_2}, which is counter-intuitive as DCNv3 is equipped with ConvNet inductive bias.

To overcome these challenges, we propose Deformable Convolution v4 (DCNv4), an innovative advancement to optimize the sparse DCN operator for practical efficiency. DCNv4 comes
with a much faster implementation and an improved operator design to enhance its performance,
which we will elaborate on as follows:

First, we conduct instruction-level kernel profiling for existing implementation and find that DCNv3 is already lightweight. 
The compute cost is less than $1\%$, while memory access costs $99\%$.
This motivates us to revisit the operator implementation and find that 
many memory accesses in the DCN forward process are redundant and thus can be optimized, leading to a much faster DCNv4 implementation.

Second, drawing inspiration from convolution's unbounded weight range, we find that softmax normalization in spatial aggregation, a standard operation in dense attention, is {\em unnecessary} in DCNv3, as it is not a requirement for operators with a dedicated aggregation window for each location. 
Intuitively, softmax puts a bounded $0\sim 1$ value range to the weight and will limit the expressive power of the aggregation weight.
This insight led us to remove the softmax in DCNv4, enhancing its dynamic property and improving its performance.

As a result, DCNv4 not only converges significantly faster than DCNv3 but also accelerates forward speed by more than $3\times$. 
This improvement allows DCNv4 to fully leverage its sparse property and become one of the fastest common core vision operators.

We further replace DCNv3 in InternImage with \ourmethod{}, creating \ournet{}.
Remarkably, \ournet{} achieves a $50\sim 80\%$ speed increase compared to InternImage without any additional modifications.
This enhancement positions \ournet{} as one of the fastest modern vision backbone networks while maintaining superior performance.
With the help of \ourmethod{}, \ournet{} significantly improves the convergence speed in ImageNet classification~\cite{deng2009imagenet} and transfer learning settings and further demonstrates improved performance in downstream tasks.

Furthermore, \ourmethod{} shows potential as a universal vision operator in various architectures and tasks. 
We integrate \ourmethod{} into other modern backbone architectures, including ConvNeXt~\cite{liu2022convnet} and ViT~\cite{dosovitskiy2020image}, replacing depthwise convolution~\cite{chollet2017xception} and dense self-attention layers~\cite{vaswani2017attention}. 
Surprisingly, without any hyperparameter adjustments, these meticulously designed networks with DCNv4 perform on par while being much faster, showing the efficacy and efficiency of the dynamic, sparse DCNv4. 
Moreover, we explore the potential of \ourmethod{} in generative models as a new application domain.
Specifically, we apply it in the U-Net~\cite{ronneberger2015u} architecture used in latent diffusion models~\cite{rombach2022high}, replacing regular convolution with DCNv4.
Our experimental results show that DCNv4 can work better than the baselines in image generation, showing great potential for using DCN to improve generative models.

We will release our implementation of DCNv4 and hope this efficient operator can facilitate future research in the vision community.

\vspace{-0.1cm}
\section{Related Work}

\paragraph{Core operators in vision models:}
The standard convolution~\cite{lecun1998gradient} stands as the most prevalent and impactful operator, forming the backbone of the majority of computer vision architectures~\cite{krizhevsky2012imagenet,simonyan2014very, he2016deep}. 
Nevertheless, a myriad of operators, each with unique characteristics, collectively play a crucial role in the development of computer vision.
Depthwise separable convolution (DWConv)~\cite{chollet2017xception} separates 
the spatial and channel operations, and has been pivotal in developing lightweight and efficient models~\cite{liu2022convnet, ma2018shufflenet}. 
RepLKNet~\cite{ding2022replknet} illustrates that a purely convolutional network, leveraging large-kernel depth-wise convolutions, can attain competitive performance in both efficiency and effectiveness. 
Deformable Convolution (DCN) series~\cite{dcnv1, dcnv2, wang2023internimage} significantly leaps the adaptability of convolution by adding learnable offsets to the convolutions kernels. Contrary to convolutions, attention mechanisms~\cite{vaswani2017attention} possess the capacity to model long-range dependencies and have been successfully adopted in various computer vision tasks ~\cite{carion2020end, dosovitskiy2020image,strudel2021segmenter, liu2022petr}. Window attention~\cite{liu2021swin,vaswani2021scaling} reduces the computational complexity inherent in vanilla attention by restricting the attention operation to a fixed-size window.
To mitigate the high computational complexity associated with vanilla attention, deformable attention~\cite{zhu2020deformable} enables each query to concentrate on a select number of key sampling points, with dynamically determined locations and weights. This efficient method is widely used in the following arts perception methods~\cite{zhang2022dino, cheng2022masked, li2023mask, li2022bevformer, yang2023bevformer, li2022panoptic}.
DynamicConv~\cite{wu2019pay} and dynamic-DWNet~\cite{han2021connection} augment depthwise convolutions (DWConv) by incorporating dynamic weights, thereby enabling the use of instance-specific weights that adapt dynamically. 
For non-grid structured data, sparse operators~\cite{thomas2019kpconv, xiong2019deformable, wang2018deep} utilize dynamic weights obtained via bilinear interpolation or in a parametric way.

\vspace{-0.4cm}
\paragraph{Memory access cost (MAC) in vision backbones:}
As underscored in previous studies~\cite{lee2019energy, ma2018shufflenet}, FLOPs, although a frequently used metric to measure model complexity, do not accurately represent the model's speed or latency. In practical scenarios, the running speed of a model is influenced by multiple factors, not just FLOPs. Memory Access Costs (MAC) play a particularly significant role in this context.~\cite{ma2018shufflenet}. 
Flash-Attention~\cite{dao2022flashattention}, by reducing the number of accesses to High Bandwidth Memory (HBM), achieves a significantly faster speed in practice despite having higher FLOPs compared to vanilla attention. 
Although DCN operators do not exhibit a disadvantage in terms of FLOPs, their latency is considerably longer compared to DW-Conv, under the same FLOPs budget, predominantly due to substantial memory access costs. 
In this work, we conduct a thorough analysis and optimization of the memory access costs associated with the DCN operators, significantly accelerating the DCN's running speed.

\vspace{-0.1cm}
\section{Method}

\begin{table}
    \centering
\resizebox{\linewidth}{!}{
    \begin{tabular}{l|rrrrrr}
      Model   & 5th EP & 10th Ep & 20th Ep & 50th Ep & 300th Ep \\
      \hline
      ConvNeXt  & 29.9 & 53.5  & 66.1 & 74.8 & 83.8\\
      \hline
      ConvNeXt & 8.5 & 25.3 & 51.1 & 69.1 & 81.6 \\
       + softmax & \textcolor{BrickRed}{(-21.4)} & \textcolor{BrickRed}{(-28.2)} & \textcolor{BrickRed}{(-15.0)} & \textcolor{BrickRed}{(-5.7)}  & \textcolor{BrickRed}{(-2.2)} \\
    \end{tabular}
}
    \vspace{-0.3cm}
    \caption{{\bf ImageNet-1K accuracy at different training epochs.} Adding softmax normalization on the convolution weights significantly affects the convergence speed and the final performance for the ConvNeXt model.}
    \label{tab:convnext_softmax}
    \vspace{-0.6cm}
\end{table}

\label{sec:convergence_speed}
\subsection{Rethinking the Dynamic Property in Deformable Convolution}

\begin{figure}
    \centering
    \includegraphics[width=\linewidth]{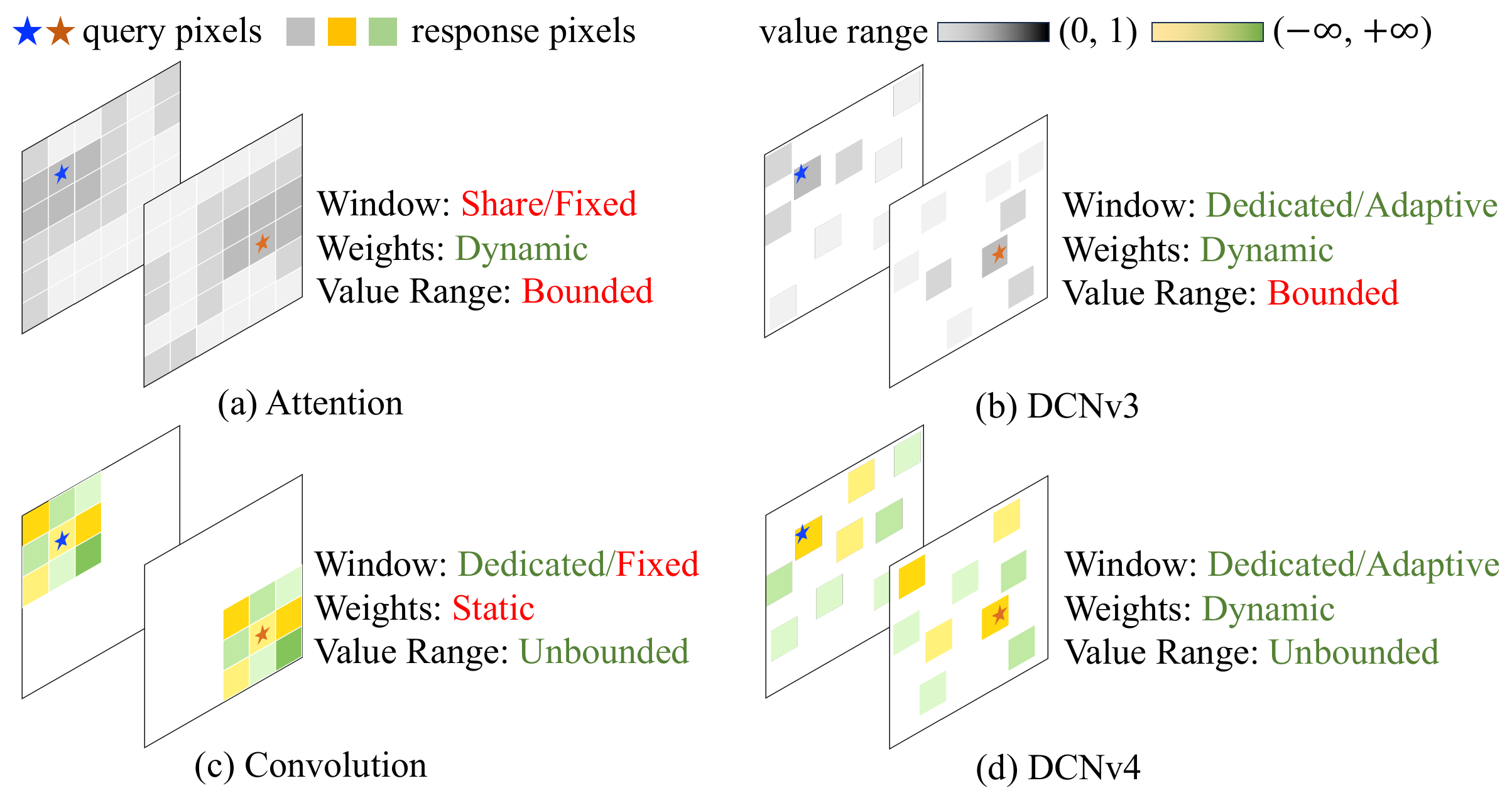}
    \vspace{-0.6cm}
    \caption{
    \textbf{Comparisons of core operators in spatial aggregation for query pixels on different locations within the same channel.} 
    (a) Attention and (b) DCNv3 use bounded (range from $0\sim 1$) dynamic weights to aggregate spatial features, while the window (sampling point set) for attention is the same, and DCNv3 uses a dedicated window for each location. (c) Convolution has a more flexible unbounded value range for aggregation weights and uses a dedicated sliding window for each location, but the window shape and aggregation weights are input-independent. (d) \ourmethod{} combines their advantages, using an adaptive aggregation window and dynamic aggregation weights with an unbounded value range.    
    }
    \label{fig:op_dynamic}
    \vspace{-0.5cm}
\end{figure}

\paragraph{Revisiting DCNv3:} Given an input $\mathbf{x}\!\in\!\mathbb{R}^{H\times W\times C}$ with height $H$, width $W$ and channel $C$, the DCNv3 operation with $K$ points is defined in Eq.~\eqref{eq:dcnv3} for each point $p_0$:
\begin{align}
    \textbf{y}_g &= \sum^{K}_{k=1} \mathbf{m}_{gk}\mathbf{x}_g(p_0 + p_k + \Delta p_{gk}), \\
    \textbf{y} &= \mathrm{concat}([\textbf{y}_1, \textbf{y}_2, ..., \textbf{y}_G], \text{axis=-1}),
    \label{eq:dcnv3}
\end{align}
where $G$ denotes the number of spatial aggregation groups.
For the $g$-th group, 
$\mathbf{x}_g, \mathbf{y}_g \in\mathbb{R}^{H \times W \times C'}$ represents the sliced input/output feature map with $C'\!=\!C/G$ represents the group dimension; 
$\mathbf{m}_{gk}\in\mathbb{R}$ denotes the spatial aggregation weights (also known as modulation scalar) of the $k$-th sampling point in the $g$-th group, conditioned on the input $\mathbf{x}$ and normalized by the softmax function along the dimension $K$; 
$p_k$ denotes the $k$-th location of the pre-defined grid sampling $\{(-1, -1), (-1, 0), ..., (0, +1), ..., (+1, +1)\}$ as in regular convolutions and
$\Delta p_{gk}$ is the offset corresponding to the grid sampling location $p_k$ in the $g$-th group. A $1\times 1$ point-wise convolution on $\mathbf{x}$ and $\mathbf{y}$ can be applied before and after the DCNv3 operator to enhance the model's expressive power, following separable convolution~\cite{chollet2017xception}.
DCNv3 is a combination of convolution and attention: on the one hand, it processes the input data in a sliding window manner, which follows convolution and inherent its inductive bias; on the other hand, the sampling offset $\Delta p$ and spatial aggregation weight $\mathbf{m}$ are dynamically predicted from the input feature, showing its dynamic property and making it more like an attention mechanism.
We compare different operators where each has its own property, as illustrated in Fig.~\ref{fig:op_dynamic}

\vspace{-0.4cm}
\paragraph{Softmax normalization:} A key difference between convolution and DCNv3 is that DCNv3 {\em normalizes} $\mathbf{m}$, the spatial aggregation weights, with a softmax function, following the convention of scaled dot-product self-attention. Conversely, convolution does not employ softmax over its weights and still works well. 
The reason why attention needs a softmax is straightforward: scaled dot-product self-attention with $Q,K,V\in\mathbb{R}^{N\times d}$ is defined with a formulation:
\begin{equation}
\mathrm{softmax}(\frac{1}{\sqrt{d}}QK^\top)V,
\label{eq:attn}
\end{equation}
where $N$ is the number of points in the same attention window (can be either global~\cite{dosovitskiy2020image} or local~\cite{liu2021swin}), $d$ is the hidden dimension, $Q, K, V$ are the query, key, and value matrices computed from the input. Softmax operation is required in Eq.~\eqref{eq:attn} for attention; without softmax, $K^\top V \in \mathbb{R}^{d\times d}$ can be calculated first, and it degrades to a linear projection for all queries in the same attention window, resulting in degenerated performance. However, for convolutional operators like depthwise convolution and DCNv3 where each point has its own dedicated aggregation window and the values in each aggregation window are already different and there is no ``key'' concept, such degradation issue no longer exists, and the normalization becomes unnecessary. In fact, normalizing convolution weights within a fixed 0-1 range using softmax could impose a significant limitation on the operator's expressive power and make the learning slower.

To confirm this hypothesis, we train a ConvNeXt model and apply softmax to the $7\times 7$ window of the depthwise convolution weights before convolution forward. We observe a remarkable decline in model performance as well as convergence speed from results in Tab.~\ref{tab:convnext_softmax}. This suggests that for operators with a dedicated aggregation window on each location like convolution or DCN, aggregation weights with an unbounded range offer better expressive power than softmax-normalized, bounded-range weights.

\vspace{-0.4cm}
\paragraph{Enhancing dynamic property:}
Motivated by this observation, we remove the softmax normalization in DCNv3, transforming the modulation scalars ranging from 0 to 1 to unbounded dynamic weights similar to convolution. As shown in Fig.~\ref{fig:op_dynamic}, this alteration further amplifies the dynamic property of DCN, where other operators have certain limits, such as bounded value range (attention/DCNv3) or fixed aggregation window with input-independent aggregation weights (convolution). 
Fig.~\ref{fig:teaser_2} shows that by making this change, \ourmethod{} converges significantly faster than DCNv3 and other common operators, including convolution and attention. 
Results in Sec.~\ref{sec:exp} further showcase that \ourmethod{} works well in both pre-training and transfer learning settings.

\vspace{-0.1cm}
\label{sec:convergence_speed}
\subsection{Speeding up DCN}
\label{sec:dcn_speedup}
Theoretically, DCN, as a sparse operator with $3\times 3$ window, should act faster than other common operators that employ larger window sizes, like dense attention or $7\times 7$ depthwise convolution. 
However, we find that this is not the case, as shown in Fig.~\ref{fig:teaser_1}. 
In this subsection, we first conduct a theoretical analysis of GPU efficiency, showing a large variance in memory access cost depending on how we read the memory. 
We further perform optimization based on our observations, significantly improving the speed of DCN by saving additional memory instruction and bringing the speed advantage of being a sparse operator into reality.

\vspace{-0.35cm}
\paragraph{Theoretical analysis of GPU efficiency}

Our study begins with a theoretical examination of the DCNv3 operator's computational behavior. We employ the roofline model to evaluate its performance, focusing on theoretical FLOPs and memory access cost (MAC). For an input and output tensor of shape $(H, W, C)$, the DCNv3 operator requires $36HWC$ FLOPs, where $3 \times 3$ represents the convolution kernel's spatial dimensions and the factor of 4 accounts for the bilinear interpolation at each sampling point.

Following the framework outlined in~\cite{ma2018shufflenet}, DCNv3's MAC is calculated as $2HWC + 27HWG$. The first term corresponds to the input/output feature map size and the second to the DCNv3's offset and aggregation weights with $G$ groups. We approximate $G$ as $C/16$ assuming a group dimsion of 16, resulting in approximately $3.7HWC$ MAC. However, this assumes an ideal scenario of infinite cache and a single memory read for each value, which is often unrealistic in parallel computing environments where concurrent thread execution necessitates simultaneous data access.

To estimate the maximum memory access requirement, we consider a scenario devoid of cache, where each output location requires fresh memory reads and involves $36$ reads for bilinear interpolation, $27$ for offset/aggregation weights, and one write operation, resulting in a MAC of $64HWC$. This is 17 times larger than the ideal case.

This analysis reveals a substantial gap in the ratio of computation-to-memory access (ranging from $0.6$ to $9.7$), highlighting the significant potential for memory access optimization. 
Notably, despite DCNv3's use of input-dependent, dynamic offsets causing non-deterministic memory access, one deterministic thing is that channels within the same group share offset values. This leads us to propose a specific optimization strategy for enhancing DCNv3's speed.

\begin{figure}
    \centering
    \includegraphics[width=\linewidth]{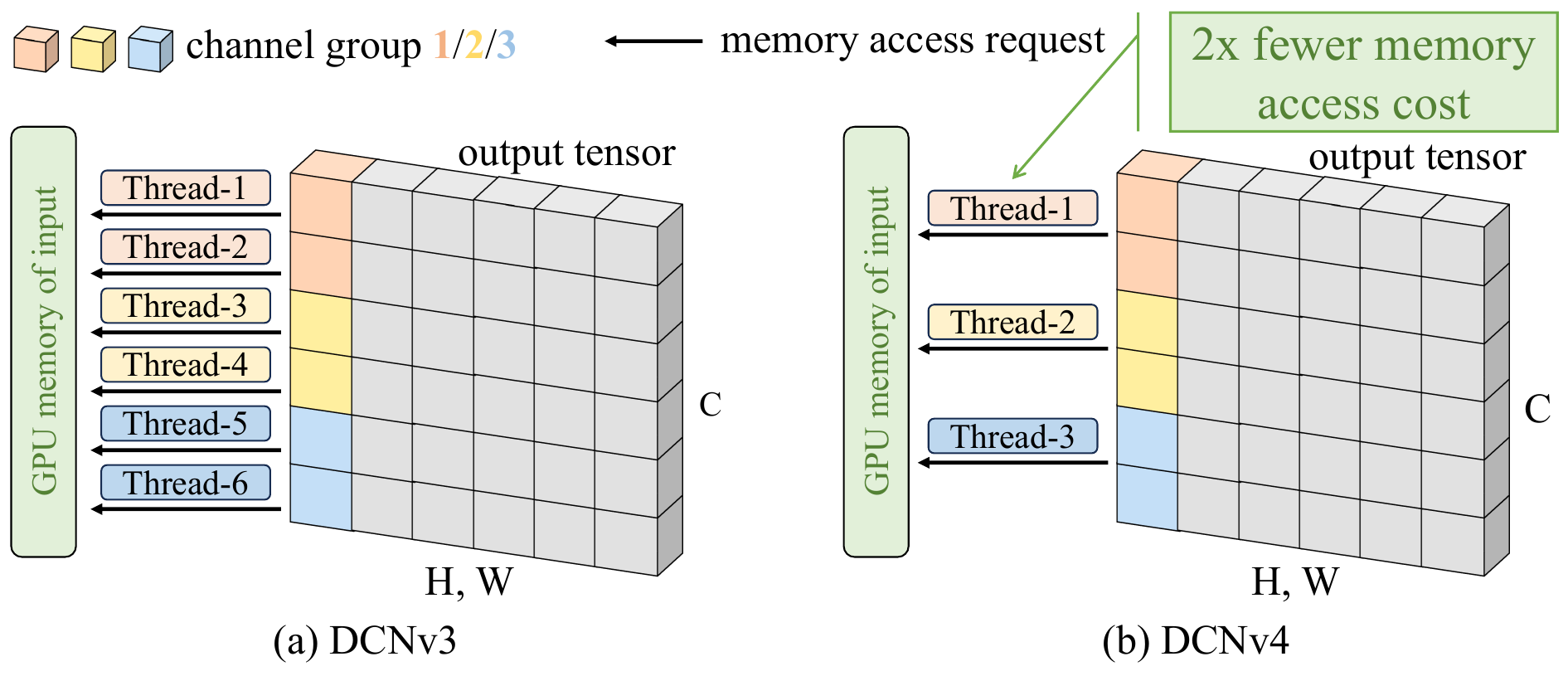}
    \vspace{-0.6cm}
    \caption{
    \textbf{Illustration of our optimization.} 
    In DCNv4, we use one thread to process multiple channels in the same group that shares sampling offset and aggregation weights. 
    Workloads like memory reading and bilinear interpolation coefficient computation can be reduced, and multiple memory access instructions can be merged.
    }
    \label{fig:op_optimization}
    \vspace{-0.4cm}    
\end{figure}

\begin{table*}
    \centering
\resizebox{0.9\linewidth}{!}{
    \begin{tabular}{l|r|r|r|r|r}
    \toprule
    \multicolumn{1}{c|}{\multirow{2}{*}{Operator}} & \multicolumn{5}{c}{Runtime (ms)} \\ 
     & $56 \times 56 \times 128$ & $28 \times 28 \times 256$ & $14 \times 14 \times 512$ & $7\times7\times1024$ & $14 \times 14 \times 768$ \\
    \hline    
      Attention (torch) & 30.8 / ~~19.3 & 3.35 / ~~2.12 & ~0.539 / 0.448 & 0.446 / ~~0.121 & 0.779 / 0.654 \\
      FlashAttention-2 & N/A / ~~2.46 & N/A / 0.451  & N/A / {\bf 0.123}  & N/A / 0.0901 & N/A / 0.163 \\
      Window Attn ($7\times 7$) & 4.05 / ~~1.46 & 2.07 / 0.770 & 1.08 / 0.422 & 0.577 / ~~0.239 & 1.58 / 0.604  \\
      DWConv ($7\times 7$, torch) & 2.02 / ~~1.98 & 1.03 / ~~1.00 & 0.515 / 0.523 & 0.269 / ~~0.261 & 0.779 / 0.773 \\
      DWConv ($7\times 7$, cuDNN)  &  0.981 / 0.438 &  0.522 / 0.267 & 0.287 / 0.153 &  0.199 / ~~0.102 & 0.413 / 0.210 \\
      DCNv3   & 1.45 / ~~1.52 & 0.688 / 0.711 & 0.294 / 0.298 & 0.125 / ~~0.126 & 0.528 / 0.548 \\
      \ourmethod{} & {\bf 0.606} / {\bf 0.404} & {\bf 0.303} / {\bf 0.230} & {\bf 0.145} / {\bf 0.123} & {\bf 0.0730} / {\bf 0.0680} & {\bf 0.224} / {\bf 0.147}\\
      \bottomrule
    \end{tabular}
}
\vspace{-0.3cm}
    \caption{{\bf Op-level benchmark on standard input shape with various downsample rates.} FP32/FP16 results are reported when the implementation is available. Our new DCNv4 can surpass all other commonly used operators under different input resolutions.
    }
    \label{tab:op_low_res}
    
\end{table*}

\vspace{-0.4cm}
\paragraph{Eliminating redundant workload:} 
In previous CUDA implementations of DCN kernel, for input with shape $(H, W, C)$\footnote{We assume the batch size is one and the memory layout is channel-last for simplicity}, offset $(H, W, G, K^2\times 2)$ and aggregation weight $(H, W, G, K^2)$, we will create $H\times W \times C$ threads in total to maximize parallelism, where each thread processes one channel for one output location. 
Notably, the $D=C/G$ channels within each group share the same sampling offset and aggregation weight values for each output location. 
Using multiple threads to process these $D$ channels on the same output location is wasteful, as different threads will read the same sampling offset and aggregation weight values from GPU memory multiple times, which is critical for a memory-bound operator. 
Processing multiple channels within the same group on each output location with one thread can eliminate these redundant memory read requests, greatly reducing memory bandwidth usage.
As the sampling locations are the same, we can also only calculate the bilinear interpolation coefficient used in DCN once.
Specifically, if each thread processes $D'$ channels, the memory access cost for reading offset and aggregation weight, as well as the computation cost for calculating bilinear interpolation coefficient, can both be reduced $D'$ times.

\vspace{-0.3cm}
\paragraph{Eliminating redundant memory instructions:} In practice, solely reusing threads for multiple channels will not see speed improvement. The reason is that when $D'$ increases, we create fewer threads and the workload of each thread now increases $D'$ times. This essentially reduces the degree of parallelism for the CUDA kernel. Luckily, the DCN kernel is now computationally lightweight as the bilinear interpolation only needs to be performed once for all $D'$ channels, and most of the workload is the memory instructions reading input values from different channels. When the memory layout is channel-last, and all $D'$ channel values are contiguous, we can leverage vectorized load: for example, to read four 32-bit float values from memory, instead of reading one 32-bit float value four times with four instructions, we can use a single instruction to load a 128-bit packed value at once, thus reducing the number of instructions and execution time of each thread. We can apply similar technique when writing the final results to GPU memory, minimizing the memory access time and increasing memory bandwidth utilization. 
Moreover, the modern half-precision data format (float16/bfloat16) halves the bytes that need to be loaded, which means the memory efficiency can be twice as much under the same memory bandwidth when using the half-precision format. However, we do not see speed improvement with half-precision data in the original DCNv3 implementation, possibly due to too much overhead on data access and computation, while in our new implementation, the speedup is significant. It is worth noting that the aforementioned optimization techniques can also be applied to DCNv1/v2 and deformable attention~\cite{zhu2020deformable}, as they share a similar performance bottleneck and issue.

\vspace{-0.4cm}
\paragraph{Micro design in DCN module: } DCNv3 module introduces multiple micro designs; as the core kernel is optimized, their impact on the speed becomes non-negligible. We identify two points in DCNv3 designs that could be further optimized: first, after removing the softmax and transforming the modulation scalar into dynamic aggregation weights as mentioned in the previous paragraph. The linear layers for computing offset and dynamic weights can actually be combined into one linear layer. This reduces network fragmentation and eliminates extra overheads, such as kernel launching and synchronization, enhancing run-time efficiency on the GPU; second, in the original DCNv3 module design, a complex sub-network that consists of depthwise $3\times 3$ conv, layer normalization (LN), GELU, and linear layer is used to compute offsets and dynamic weights. Following the design in Xception~\cite{chollet2017xception}, we remove the additional LN-GELU layers and use the original separable convolution structure, further reducing running time. We empirically find that if latency is a higher priority, the depthwise convolution can also be removed with only a minor performance sacrifice.

\begin{table*}
    \centering
\resizebox{0.9\linewidth}{!}{
    \begin{tabular}{l|r|r|r|r|r}
    \toprule
    \multicolumn{1}{c|}{\multirow{2}{*}{Operator}} & \multicolumn{5}{c}{Runtime (ms)} \\ 
     & $200 \times 320 \times 128$ & $100 \times 160 \times 256$ & $50 \times 80 \times 512$ & $25\times 40\times 1024$ & $64 \times 64 \times 768$ \\
    \hline          
      Attention (torch)   & OOM / OOM~~ & 25.4 / ~~~~12.9 & 2.88 / ~~~~1.89  & 0.490 / ~~0.309 & 4.17 / ~~~~2.57 \\
      FlashAttention-2    & N/A / ~~13.2~~ & N/A / ~~~~1.74 & N/A / ~~0.285 & N/A / 0.0797 & N/A / ~~0.437\\
      Window Attn ($7\times 7$)  & 1.33 / 0.509~~ & 0.728 / ~~0.291 & 0.426 / ~~0.186 &  0.279 / ~~0.165 & 0.673 / ~~0.272\\
      DWConv ($7\times 7$, torch) & 0.634 / 0.608~~ & 0.313 / ~~0.315 & 0.167 / ~~0.158 & 0.0943 / 0.0894 & 0.260 / ~~0.253 \\
      DWConv ($7\times 7$, cuDNN) & 0.331 / 0.282~~ & 0.188 / ~~0.168 & 0.114 / ~~0.115 & 0.0817 / 0.0881 & 0.161 / ~~0.156  \\
      DCNv3   & 0.472 / 0.493~~ & 0.244 / ~~0.253 & 0.128 / ~~0.132 & 0.0737 / 0.0767 & 0.194 / ~~0.199  \\
      \ourmethod{}   & {\bf 0.210} / {\bf 0.136}~~ & {\bf 0.124} / {\bf 0.0895} & {\bf 0.0707} / {\bf 0.0589} & {\bf 0.0452} / {\bf 0.0426} & {\bf 0.103} / {\bf 0.0672}\\
      \bottomrule
    \end{tabular}
}
    \caption{{\bf Op-level benchmark on high-resolution input shape with various downsample rates.} 
    DCNv4 performs well as a sparse operator, surpassing all other baselines, while dense global attention is slow under this scenario.
    }
    \label{tab:op_high_res}
\end{table*}

\vspace{-0.3cm}
\section{Experiments}
\label{sec:exp}

In this section, we verify the effectiveness of our proposed DCNv4 module from both speed as well as performance perspective. 
We first benchmark the operator-level speed and integrate DCNv4 into the backbone model to test the system-level performance further. 
All speed test results are obtained with an NVIDIA A100 80G SXM GPU.
Due to the space limit, we include additional experimental results and implementation details, including other hyperparameter settings and hardware/software environment, in supp.

\vspace{-0.1cm}
\subsection{Speed Benchmark for Operators}
\label{sec:op_benchmark}
\vspace{-0.1cm}
\paragraph{Settings:} 
We conduct the op-level benchmark by solely measuring the running time of several representative operators building state-of-the-art vision backbone models, including full attention~\cite{vaswani2017attention} implemented with PyTorch as well as the advanced FlashAttention-2~\cite{dao2023flashattention} implementation, window-based attention with window size $7\times 7$~\cite{liu2021swin}, depthwise convolution with $7\times 7$ window, implemented by cuDNN~\cite{chetlur2014cudnn} and ATen library from PyTorch~\cite{paszke2019pytorch}, respectively. 
For simplicity, we only benchmark the core operation for spatial aggregation, and additional linear layers like qkv projection and output projection layers are excluded and not included in the runtime measurement. Please refer to supp. for a more comprehensive module-level comparison.
For the input scale, we first consider a feature map shape generated from the standard $224\times 224$ input resolution for image classification with 4, 8, 16, 32$\times$ downsample ratio as used by common hierarchical ConvNet/transformer backbones; we also add a feature map shape from isotropic backbone like ViT with a downsampling ratio 16 and larger hidden dimension.
We further consider high-resolution inputs often used in downstream tasks like object detection. We set the input shape to be $800\times 1280$ and $1024\times 1024$ for the hierarchical feature map and isotropic feature map, respectively, as they are the common practice in object detection~\cite{he2017mask,li2022exploring}. 
Batch size is 64 and 1 for these two input sets, respectively. 
For operators with a head/group concept, we set the dimension of each head/group to 32 and change the number of heads/groups when the hidden dimension varies.

\vspace{-0.2cm}
\paragraph{Results:} We show the benchmark results on standard resolution and high-resolution input in Tab.~\ref{tab:op_low_res} and Tab.~\ref{tab:op_high_res}, respectively. 
We report results with both FP32 and FP16 data formats except FlashAttention, which does not have an FP32 implementation.
Dense global attention implemented with PyTorch performs significantly slower when the input resolution is large and even out of memory.
FlashAttention significantly improves the speed of attention and can be even faster than $7\times 7$ window attention in certain cases, indicating the importance of proper optimization. 
However, it does not change the quadratic complexity of attention; when the input resolution is high, it still falls behind local/sparse operators like window attention or convolution.
While DCNv3 can be faster than DWConv with plain implementation, it is slower than the heavily optimized cuDNN version.
Instead, our \ourmethod{} can provide more than $3\times$ speedup compared to DCNv3, greatly saving the running time. 
Moreover, \ourmethod{} can successfully leverage the advantage of using a $3\times 3$ sparse window to perform significantly faster than other baselines under different settings.

\vspace{-0.1cm}
\subsection{Image Classification}
\label{sec:backbone_benchmark}
\paragraph{Settings:} 
We evaluate the effectiveness of DCNv4 on the ImageNet classification task. 
We start from a strong baseline, InternImage~\cite{wang2023internimage}, as it shows state-of-the-art performance among ConvNet-based models.
We replace the original DCNv3 in InternImage with \ourmethod{} and create \ournet{}. 
All other implementation details, including network architecture and hyperparameters, are kept the same as~\cite{wang2023internimage}.
We also compare Swin-Transformer and ConvNeXt which are two representative baselines in Transformer and ConvNet models.
We follow the common practice~\cite{liu2021swin, liu2022convnet, wang2023internimage} and train FlashInternImage-Tiny/Small/Base on ImageNet-1K for 300 epochs. FlashInternImage-Large is first trained on ImageNet-22K for 90 epochs and then fine-tuned on ImageNet-1K for 20 epochs. Other baselines share the same setting for a fair comparison.

\begin{table}[t]
    \centering
\resizebox{0.9\linewidth}{!}{
\setlength{\tabcolsep}{3pt}
    \begin{tabular}{l|cccc}
    \toprule
    \multicolumn{1}{c|}{Model} & Size & Scale & Acc & Throughput\\
    \hline
      Swin-T   & 29M & 224$^2$ & 81.3 & 1989 / 3619 \\
      ConvNeXt-T  & 29M & 224$^2$ & 82.1 & 2485 / 4305\\
      InternImage-T  & 30M & 224$^2$ & 83.5 & 1409 / 1746\\
      \rowcolor{gray!20}
      FlashInternImage-T  & 30M & 224$^2$ & {\bf 83.6} & \begin{tabular}{@{}c@{}}2316 / 3154 \\ {\footnotesize\textcolor{ForestGreen}{($+64\%/+80\%$)}} \end{tabular}  \\
      \hline
      Swin-S   & 50M & 224$^2$ & 83.0 & 1167/2000\\
      ConvNeXt-S  & 50M & 224$^2$ & 83.1 & 1645/2538 \\
      InternImage-S  & 50M & 224$^2$ & 84.2& 1044/1321 \\
      \rowcolor{gray!20}
      FlashInternImage-S  & 50M & 224$^2$  & {\bf 84.4} & 1625 / 2396  \\
      \hline 
      Swin-B   & 88M & 224$^2$ & 83.5 &  \ 934 / 1741\\
      ConvNeXt-B   & 89M  & 224$^2$ & 83.8 & 1241 / 1888 \\
      InternImage-B   & 97M & 224$^2$ & {\bf 84.9} & \ 779 / 1030 \\
      \rowcolor{gray!20}
      FlashInternImage-B  & 97M & 224$^2$ & {\bf 84.9} & 
      \begin{tabular}{@{}c@{}}1174 / 1816 \\ {\footnotesize\textcolor{ForestGreen}{($+51\%/+76\%$)}} \end{tabular}
       \\
      \hline
      Swin-L   & 197M & 384$^2$ & 87.3 & 206 / 301\\
      ConvNeXt-L   & 198M & 384$^2$ & 87.5 & 252 / 436\\
      InternImage-L   & 223M & 384$^2$ & 87.7 & 158 / 214 \\
      ConvNeXt-XL   & 350M & 384$^2$ & 87.8 & 170 / 299\\
      InternImage-XL   & 335M & 384$^2$ & 88.0 & 125 / 174 \\
      \rowcolor{gray!20}
      FlashInternImage-L   & 223M & 384$^2$ & {\bf 88.1} & 
      \begin{tabular}{@{}c@{}}248 / 401 \\ {\footnotesize\textcolor{ForestGreen}{($+57\%/+87\%$)}} \end{tabular}
      \\
    \bottomrule
    \end{tabular}
}
\vspace{-0.2cm}
    \caption{
    {\bf Image classification performance on ImageNet-1K.} 
    We show relative speedup between FlashInternImage w/ \ourmethod{} and its InternImage counterparts. \ourmethod{} significantly improves the speed while shows state-of-the-art performance.
    }
    \label{tab:backbone_comparsion}
    \vspace{-0.4cm}
\end{table}

\vspace{-0.4cm}
\paragraph{Results:} Tab.~\ref{tab:backbone_comparsion} shows the results of models at various scales. Besides the model size and training/inference resolution, we also report each model's overall throughput (number of images per second) in FP32/FP16 data format. 
For a fair comparison, we use timm~\cite{rw2019timm} implementation of ConvNeXt and Swin Transformer, which is actually faster than the original implementation.
Equipped with \ourmethod{}, FlashInternImage significantly improves the throughput of the InternImage counterpart over $50\%\sim80\%$ and slightly improves the model performance.
FlashInternImage now matches the speed of ConvNeXt with higher accuracy. It is worth noting that FlashInternImage-S can outperform ConvNeXt-B (84.4\% \vs 83.8\%) while being faster than it, showing a better speed-accuracy trade-off. 
Moreover, the FlashInternImage-L can even surpass ConvNeXt-XL and InternImage-XL and being $30\%\sim 130\%$ (401 \vs 174) faster, showing the effectiveness of our \ourmethod{} module.

\subsection{Downstream Tasks with High-Resolution Input}
\label{sec:downstream_benchmark}
We further evaluate the performance of \ourmethod{} on representative downstream perception tasks with high-resolution input, including instance segmentation, semantic segmentation and 3D object detection. 
We keep all implementation details the same as InternImage and only change the backbone model to FlashInternImage for a fair comparison. 
The backbone models are initialized from the ImageNet pretrained weights when training the downstream models.

\begin{table}
    \centering
\resizebox{\linewidth}{!}{
\setlength{\tabcolsep}{3pt}
    \begin{tabular}{l|cr|cc|cc}
\toprule
        \multirow{3}{*}{Model} & \multirow{3}{*}{\#param} & \multicolumn{1}{c|}{\multirow{3}{*}{\begin{tabular}{@{}c}
        FPS
    	\end{tabular}}} &
        \multicolumn{4}{c}{Mask R-CNN}
        \\
         ~ & ~ & ~ &  \multicolumn{2}{c}{1$\times$} & \multicolumn{2}{c}{3$\times$+MS}\\
        \cline{4-7} ~ & ~ & ~ & $\rm AP^b$ & $\rm AP^m$ & $\rm AP^b$ & $\rm AP^m$  \\

     \hline

      Swin-T & 48M &  66 / 106 & 42.7 & 39.3 &  46.0 &  41.6\\
      ConvNeXt-T   & 48M & 78 / 113 & 44.2 & 40.1 & 46.2 & 41.7 \\
      InternImage-T   & 49M & 54 / ~~69 &  47.2 & 42.5 & 49.1 & 43.7 \\
      \rowcolor{gray!20}
      FlashInternImage-T  & 49M & 72 / 102 & {\bf 48.0} & {\bf 43.1} & {\bf 49.5} & {\bf 44.0} \\
      \hline
      Swin-S   & 69M & 45 / ~~77 & 44.8 & 40.9 & 48.2 & 43.2  \\
      ConvNeXt-S   & 70M & 54 / ~~83 & 45.4 & 41.8 & 47.9 & 42.9\\
      InternImage-S   & 69M & 44 / ~~56 & 47.8 & 43.3 & 49.7 & 44.5 \\
      \rowcolor{gray!20}
      FlashInternImage-S & 69M & 57 / ~~83  & {\bf 49.2} & {\bf 44.0} & {\bf 50.5} & {\bf 44.9} \\ 
      \hline 
      Swin-B   & 107M & 33 / ~~59 & 46.9 & 42.3 & 48.6 & 43.3 \\
      ConvNeXt-B   & 108M & 43 / ~~70 & 47.0 & 42.7 & 48.5 & 43.5\\
      InternImage-B   & 115M & 33 / ~~43 & 48.8 & 44.0 & 50.3 & 44.8 \\
      \rowcolor{gray!20}
      FlashInternImage-B  & 115M & 44 / ~~67 & {\bf 50.1} & {\bf 44.5} & {\bf 50.6} & {\bf 45.4} \\
      \multicolumn{7}{c}{ } \\
        \multirow{3}{*}{Model} & \multirow{3}{*}{{\#param}} & \multicolumn{1}{c|}{\multirow{3}{*}{\begin{tabular}{@{}c}
        FPS 
    	\end{tabular}}} &
        \multicolumn{4}{c}{Cascade Mask R-CNN}
        \\
        ~ & ~ & ~ &  \multicolumn{2}{c}{1$\times$} & \multicolumn{2}{c}{3$\times$+MS}\\
        \cline{4-7} ~ & ~ & ~ &  $\rm AP^b$ & $\rm AP^m$ & $\rm AP^b$ & $\rm AP^m$  \\
    
     \hline
      Swin-L   & 253M & 20 / ~~26 & 51.8 & 44.9 & 53.9 & 46.7 \\
      ConvNeXt-L   & 255M & 26 / ~~40 & 53.5 & 46.4 & 54.8 & 47.6  \\
      InternImage-L  & 277M & 20 / ~~26 & 54.9 & 47.7 & 56.1 & 48.5 \\
      ConvNeXt-XL   & 407M & 21 / ~~32 & 53.6 & 46.5 & 55.2 & 47.7 \\ 
      InternImage-XL  & 387M & 16 / ~~23 & 55.3 & 48.1 & 56.2 & 48.8 \\
      \rowcolor{gray!20}
      FlashInternImage-L   & 277M & 26 / ~~39 & {\bf 55.6} & {\bf 48.2} & {\bf 56.7} & {\bf 48.9} \\
\bottomrule
    \end{tabular}
}
\vspace{-0.2cm}
    \caption{{\bf Object detection and instance segmentation performance on COCO val2017.} $\rm AP^b$ and $\rm AP^m$ denotes box AP and mask AP, respectively. ``MS'' means multi-scale training. FlashInternImage w/ \ourmethod{} models converge faster, clearly outperform other baselines with $1\times$ training schedule, and still maintain a leading position when training $3\times$ longer while being significantly faster than InternImage. 
}
    \label{tab:detection}
\vspace{-0.4cm}
\end{table}

\paragraph{Instance Segmentation:} 
We train FlashInternImage with two representative instance segmentation frameworks, Mask R-CNN~\cite{he2017mask} and Cascade Mask-RCNN~\cite{cai2018cascade}, on COCO dataset~\cite{lin2014microsoft} at 1$\times$ (12 epochs) and 3$\times$ (36 epochs) training schedules. 
The results are shown in Tab.~\ref{tab:detection}.
We also report FPS with batch size 16 in FP32/FP16 data format.
\ournet{} shows superior results on all model scales and training schedules, achieving a higher speed-accuracy tradeoff.
For example, FlashInternImage-T/S surpasses all other models with the same scale and is on par with a larger InternImage-S/B while being $80\%-90\%$ faster.

\paragraph{Semantic Segmentation:} 
We train FlashInternImage with UperNet~\cite{xiao2018unified} on ADE20K~\cite{zhou2017scene} dataset for 160K iterations. 
We can draw a similar conclusion as instance segmentation from the results in Tab.~\ref{tab:segmentation}, with FPS reported with batch size 16 in FP32/FP16. FlashInternImage w/ DCNv4 can achieve significantly faster speed and further improve the performance of InternImage across different model scales, resulting in a new state-of-the-art.

\paragraph{3D Detection:} 
We further test \ourmethod{} on the camera-based 3D object detection task in the autonomous driving scenario, 
We train BEVFormer v2~\cite{yang2023bevformer}, a state-of-the-art multi-camera 3D object detector, with FlashInternImage-Small and Base backbone models on the nuScenes dataset for 24 epochs. 
We report results on the nuScenes test set in Tab.~\ref{tab:bevformer} with FPS for each model. We note that the header parts, such as the BEV encoder and object decoder in BEVFormer v2, are {\em underoptimized} and take more than 50\% of the running time (and even more with a faster backbone); thus, we also report the FPS for the backbone for a clearer illustration. 
Our results show that when only considering the backbone, FlashInternImage can be twice or even three times faster than the InternImage backbone with an on-par performance, greatly increasing the model efficiency.

\begin{table}
    \centering
\resizebox{\linewidth}{!}{
\setlength{\tabcolsep}{3pt}
    \begin{tabular}{l|c|cr|cc}
    \toprule
    	\multirow{2}{*}{Model} & crop & \multirow{2}{*}{\#param} & \multirow{2}{*}{\begin{tabular}{@{}c}
     FPS
    	\end{tabular}} & mIoU & mIoU\\
    	& size & & & (SS) & (MS)   \\
    	\hline
      Swin-T & $512^2$ & 60M & 107 / 168& 44.5 & 45.8 \\
      ConvNeXt-T & $512^2$ & 60M& 120 / 184 & 46.0 & 46.7 \\
      InternImage-T &$512^2$ &  59M & 100 / 139 & 47.9 & 48.1 \\
      \rowcolor{gray!20}
      FlashInternImage-T & $512^2$ & 59M & 119 / 206 & {\bf 49.3} & {\bf 50.3} \\
      \hline 
      Swin-S  &$512^2 $&  81M & 89 / 142 & 47.6 & 49.5 \\
      ConvNeXt-S  & $512^2 $& 82M & 107 / 164 & 48.7 & 49.6\\
      InternImage-S &$512^2 $&  80M  & 89 / 123 & 50.1 & 50.9  \\
      \rowcolor{gray!20}
      FlashInternImage-S & $512^2 $& 80M & 107 / 182 & {\bf 50.6} & {\bf 51.6} \\ 
      \hline
      Swin-B  & $512^2 $& 121M & 77 / 126 & 48.1 & 49.7\\
      ConvNeXt-B & $512^2 $& 122M  & 95 / 147 & 49.1 & 49.9\\
      InternImage-B & $512^2 $& 128M  & 77 / 104 & 50.8 & 51.3 \\
      \rowcolor{gray!20}
      FlashInternImage-B & $512^2$& 128M & 94 / 157 & {\bf 52.0} & {\bf 52.6}\\
      \hline
      Swin-L & $640^2$ & 234M  & 59 /\ \ \  99 & 52.1 & 53.5 \\
      ConvNeXt-L  &$640^2$ & 235M &  73 / 117 & 53.2 & 53.7 \\
      InternImage-L & $640^2$ & 256M & 56 /\ \ \  78 & 53.9 & 54.1 \\
      ConvNeXt-XL & $640^2$ & 391M & 53 /\ \ \  75 & 53.6 & 54.0  \\
      InternImage-XL & $640^2$ & 368M & 47 /\ \ \  67 & 55.0 & 55.3 \\
      \rowcolor{gray!20}
      FlashInternImage-L  & $640^2$ & 256M & 71 / 122 & {\bf 55.6} & {\bf 56.0} \\
    \bottomrule
    \end{tabular}
}
    \vspace{-0.15cm}
    \caption{{\bf Semantic segmentation performance on the ADE20K validation set.} All models are trained with UperNet. ``SS'' and ``MS'' denote single-scale and multi-scale testing, respectively. FPS is reported with single-scale testing. FlashInternImage w/ DCNv4 achieves superior performance with competitive speed.}
    \label{tab:segmentation}
    \vspace{-0.35cm}
\end{table}

\begin{table}
    \centering
\resizebox{\linewidth}{!}{
    \begin{tabular}{l|cccc}
    \toprule
    Model  & NDS & mAP & FPS$^\dag$ & \textcolor{gray}{FPS}\\
    \hline
      InternImage-B   & 62.0 & 54.0 & 8.0 & \textcolor{gray}{2.7}\\
      InternImage-XL   & 63.4 & 55.6 & 4.0& \textcolor{gray}{2.0}\\  %
      \hline
      FlashInternImage-S   & 61.7 & 55.5 & 16.8 & \textcolor{gray}{4.1} \\
      FlashInternImage-B   & 63.1 & 57.4 & 12.1 & \textcolor{gray}{3.8} \\   
      \bottomrule
    \end{tabular}
}
    \vspace{-0.2cm}
    \caption{{\bf 3D detection performance of BEVFormer v2 on nuScenes {\em test} set.} 
    We report FPS for the backbone (denoted with $\dag$), \textcolor{gray}{overall FPS} results with {\em underoptimized} head implementation are also added for reference.
    With on-par NDS and higher mAP results, FlashInternImage can be $50-90\%$ faster than InternImage baselines or $200\%-300\%$ when only considering the backbone.
    }
    \vspace{-0.3cm}

    \label{tab:bevformer}
\end{table}

\subsection{\ourmethod{} as a Universal Operator}
\label{sec:universal_op}

\vspace{-0.1cm}
\paragraph{Drop-in replacement in other vision backbones :} 
We verify whether \ourmethod{} can still work well in architectures designed with other operators, such as ConvNeXt and ViT.
To achieve that, we replace the attention module and depthwise convolution layer with \ourmethod{} in ViT and ConvNeXt and perform supervised learning on ImageNet-1K without changing any other architecture and hyperparameters, similar to FlashInternImage and InternImage.
The results are shown in Tab.~\ref{tab:dropin_convnext_vit}. 
We can see that on these architectures, which are carefully tuned for the specific operators, our \ourmethod{} can perform equally well. 
Thanks to the fast speed of \ourmethod{}, the new model can even achieve better throughput, showcasing the superior performance of \ourmethod{}.

\begin{table}
    \centering
    \begin{tabular}{l|cl}
    \toprule
    Model & IN-1K Acc & Throughput \\
    \hline
       ConvNeXt-B & 83.8 & 1241 / 1888 \\
      \rowcolor{gray!20}
       ConvNeXt-B + \ourmethod{}  & 84.0 & \begin{tabular}{@{}l}
        1495 / 2513 \\
         {\footnotesize\textcolor{ForestGreen}{($+20\%/+33\%$)}}
       \end{tabular}\\
       \hline
       ViT-B & 81.8 & 1683 / 2781$^\dag$ \\
      \rowcolor{gray!20}
       ViT-B + \ourmethod{} & 81.9 & 
       \begin{tabular}{@{}l}
       2092 / 3261 \\
         {\footnotesize\textcolor{ForestGreen}{($+24\%/+17\%$)}}       
       \end{tabular}\\
        \bottomrule
    \end{tabular}
    \vspace{-0.2cm}
    \caption{\textbf{\ourmethod{} in other architecture.} We show supervised learning results on ImageNet-1K and throughput. \ourmethod{} achieves higher throughput with comparable accuracy. $\dag$ denotes testing with the advanced FlashAttention-2 implementation.}
    \label{tab:dropin_convnext_vit}
    \vspace{-0.35cm}
\end{table}

\vspace{-0.3cm}
\paragraph{Drop-in replacement in diffusion model:} 
DCN has been recognized to be an effective operator for perception tasks. 
As generative models become a fundamental tool for AI-generated content (AIGC), we are also curious if \ourmethod{} can work well on generation tasks with diffusion-based generative models. 
Specifically, we choose the 
U-Net~\cite{ronneberger2015u} used in Stable Diffusion~\cite{rombach2022high} as our baselines and replace the attention module and regular $3\times 3$ convolution in 
U-Net.
We use U-ViT's codebase, follow its training schedule, and train a latent diffusion model based on the image latent extracted from an image autoencoder provided by Stable Diffusion. 
We show the results in Tab.~\ref{tab:dropin_diffusion}. We can see that DCNv4 also works well in generative modeling, achieving better results in terms of FID/Throughput with fewer parameters compared to regular convolution in U-Net. 
Notice that the architecture/hyperparameters may not be optimal for DCNv4, and it is possible that re-designing the models or searching for new hyperparameters for DCNv4 will give better results.

\begin{table}
    \centering
    \begin{tabular}{l|ccc}
    \toprule
        Model & \#param & FID & FPS \\
        \hline
         U-Net & 860M & 2.94 & 4.82 \\
      \rowcolor{gray!20}
         U-Net + DCNv4 & 566M  & 2.44 & 4.92 \\ %
        \bottomrule
    \end{tabular}
    \vspace{-0.2cm}
    \caption{\textbf{Class conditional generation on ImageNet 256x256.} Latent diffusion models are trained from scratch with U-Net. 
    We replace the convolution layer in the models with DCNv4. 
    DCNv4 can achieve better FID results without any hyperparameter tuning. 
    }
    \label{tab:dropin_diffusion}
    \vspace{-0.2cm}
\end{table}

\begin{table}
    \centering
\resizebox{0.85\linewidth}{!}{
    \begin{tabular}{l|rr}
    \toprule
    Implementation variant & Module & Kernel \\
    \hline
       Original DCNv3 & 3.28 & 1.45 \\
   \hline
       - micro design & 2.12 & 1.45 \\
       - redundant memory access  & 2.20 & 1.53 \\
       - redundant computation  & 2.18 & 1.51 \\
       \rowcolor{gray!20}
       - redundant memory instr. & 1.28 & 0.606\\
       \rowcolor{gray!20}
       - half-precision format & 0.873 & 0.404\\
    \bottomrule
    \end{tabular}
}
\vspace{-0.2cm}
    \caption{\textbf{Ablation studies of DCN's runtime (ms).} 
    We show how to achieve DCNv4 (\textcolor{gray}{gray} row) from the original DCNv3 implementation and how different design choices affect the speed.
    }
    
    \label{tab:ablation}
    \vspace{-0.4cm}
\end{table}

\vspace{-0.1cm}
\subsection{Ablation Study}
\label{sec:ablation}
\vspace{-0.1cm}

We conduct ablation studies in our optimization choice described in Sec.~\ref{sec:dcn_speedup}. The results are shown in Tab.~\ref{tab:ablation}. 
The time in the table is obtained with $56\times 56\times 128$ input with batch size 64 and 4 groups (32 channels per group). 
We first remove the softmax operation and improve the micro design, which means we merge the two linear layers into one and remove costly layer norm and GELU activation in offset/aggregation weight computing, simplifying the overall modules and increasing the speed.
We then start modifying the kernel implementation. 
First, we change the parallel execution pattern and let each thread process 8 channels instead of 1 channel, thus, unnecessary memory access on loading sampling offset and aggregation weight values from the GPU memory can be saved. 
As we expected, solely applying this change will not increase the speed as the degree of parallelism decreases, and each thread's workload increases 8 times now.
The latency is increased instead.
Eliminating redundant computation by reusing the bilinear interpolation coefficient (4th row) can save some time but is insignificant.
Removing the redundant memory instruction via vectorized load/store can greatly reduce the workload of each thread and largely accelerate the GPU kernel speed (5th row).
Using a half-precision datatype, which halves the number of bytes the kernel needs to read/write, further increases the data throughput, as shown in the 6th row. 
In the end, we reach the final DCNv4 design, which is three times more efficient than the original implementation.

\vspace{-0.1cm}
\section{Conclusion}
\vspace{-0.2cm}

We present Deformable Convolution v4 (DCNv4), an efficient dynamic and sparse operator.
By rethinking the dynamic property in deformable convolution and streamlining memory access, DCNv4 emerges as a much faster and more effective operator than its predecessor DCNv3. 
DCNv4-equipped FlashInternImage backbone not only enhances speed but also improves performance across various vision tasks. 
We further show DCNv4's versatility and effectiveness as a universal operator by integrating it into state-of-the-art architecture like ConvNeXt and ViT with improved throughput and accuracy; and it also works well in latent diffusion model, showing its potential to enhance generative models.

\appendix

\setcounter{figure}{0}
\setcounter{table}{0}
\maketitlesupplementary

\section{Implementation Details}

\paragraph{Environment:}  We use an A100 80GB SXM GPU to benchmark throughput on all experiments. The software environment is PyTorch 1.13, CUDA 11.7, cuDNN 8.5. When testing Flash Attention~\cite{dao2023flashattention}, we use Flash Attention 2.3.1. When testing Window Attention in Swin Transformer, we use the PyTorch implementation from timm 0.9.7~\cite{rw2019timm}.

\paragraph{2D object detection on COCO:}
To validate the effectiveness of our method in 2D object detection, we employed two object detection methods: Mask R-CNN and Cascade Mask R-CNN, primarily referring to the settings of Internimage. Following common practices. We used two schedules: 1x (12 epochs) and 3x (36 epochs) to respectively assess the convergence speed and final performance of our model. For the 1$\times$ schedule, images are resized such that the shorter side is 800 pixels, with the longer side not exceeding 1,333 pixels. During the evaluation phase, the shorter side of input images is consistently set to 800 pixels. For the 3$\times$ schedule, the shorter side is resized to a range between 480 and 800 pixels, while the longer side remains capped at 1,333 pixels. The base learning rate is set at 1e-4 for a batch size of 16. We employ the AdamW optimizer, incorporating a weight decay of 0.05. The initialization of the backbone is the pre-trained classification weights.

\paragraph{2D semantic detection on ADE20K:}

We employed the UperNet to validate the effectiveness of our method in 2D semantic segmentation on the ADE20K dataset. Our experimental setup is primarily based on InternImage. For FlashInternImage-T/S/B and FlashInternImage-L, we use the AdamW optimizer with learning rates of 6e-5 and 2e-5, respectively. The crop size for FlashInternimage T/S/B is set to 512, while for FlashInternImage-L, it is set to 640.
We train all our models using a batch size of 16 for 160k iterations to ensure a fair comparison with previous methods. The initialization of the backbone is also the pre-trained classification weights.

\paragraph{3D object detection on nuScenes:}

We employed BEVFormerV2 to validate the effectiveness of our method in 3D object detection on nuScenes. Adhering to the settings of BEVFormerV2, the backbone initialization is pretrained on the COCO dataset.  In alignment with BEVFormerV2, we utilized data spanning a total of 8 seconds, encompassing both past and future information. We use the AdamW optimizer with a batch size of 16 and a learning rate of 4e-4. We train the model for 24 epochs.

\section{Additional Experimental Results}

\paragraph{Downstream results with advanced headers:}
We employed the more advanced DINO~\cite{zhang2022dino} and Mask2Former~\cite{cheng2022masked} to further validate the effectiveness of our FlashInternImage, as shown in \cref{tab:detection_app} and \cref{tab:segmentation_app}. For 2D object detection on COCO with DINO head, we train our model with 12 epochs. During the training phase, we adopt the same multi-scale training strategy that introduced above. Other settings, including optimizer, weight decay, and learning rate, are also the same as those used in Mask-RCNN. For 2D semantic segmentation on ADE20K, we set the learning rate to be 1e-4 with a batch size of 16. For Base and Large scale, we use a crop size of 640.
Other settings are the same as those used in UperNet. 

Under the application of more advanced task heads, our method still maintains a significant advantage in accuracy while also offering competitive inference speed.

\begin{table}
    \centering
\resizebox{\linewidth}{!}{
\setlength{\tabcolsep}{3pt}
    \begin{tabular}{l|cr|cccc}
\toprule
        \multirow{2}{*}{Model} & \multirow{2}{*}{\#param} & \multicolumn{1}{c|}{\multirow{2}{*}{\begin{tabular}{@{}c}
        FPS
    	\end{tabular}}} &
        \multicolumn{4}{c}{DINO}
        \\
        \cline{4-7} ~ & ~ & ~ & $\rm AP$ & $\rm AP^S$ & $\rm AP^M$ & $\rm AP^L$  \\

     \hline

      Swin-T &  48.2M& 37 / 50 & 51.1 & 34.3 &  53.9 &  66.1\\   ConvNext-T & 48.5M & 39 / 54 & 51.0 & 33.5 &  53.9 &  65.8\\
      InternImage-T  & 48.7M&33 / 37  &  53.9 & 37.7 & 57.7 & 68.7 \\
      \rowcolor{gray!20}
      FlashInternImage-T  & 48.6M & 39 / 48 & {\bf 54.7} & {\bf 37.9} & {\bf 58.6} & {\bf 69.8} \\
      \hline
      Swin-B   & 108M & 24 / 32 & 53.1 & 36.8 & 56.7 & 68.9  \\
      ConvNeXt-B   & 109M & 28 / 36 & 53.1 & 35.5 & 56.6 & 68.5\\
      InternImage-B   & 116M & 24 / 26 & 54.8 & 38.2 & 58.6 & 70.3 \\
     \rowcolor{gray!20}
     FlashInternImage-S & 68.8M &  33 / 41 & {55.3} & {39.0} & {59.2} & {71.1} \\ 
      \rowcolor{gray!20} 
      FlashInternImage-B & 116M & 28 / 35  & {\bf 56.0} & {\bf 41.2} & {\bf 59.8} & {\bf 71.2} \\       
      \hline 
      Swin-L   & 218M & 17 / 24 & 56.1 & 39.6 & 59.8 & 71.6 \\
      ConvNeXt-L   & 219M & 20 / 27 &55.9 & 39.8 & 59.5 & 71.1  \\
      InternImage-L   & 241M &17 / 18  & 57.6 & {\bf 44.1} & 61.5 & 73.4 \\
      \rowcolor{gray!20}
      FlashInternImage-L  & 241M & 20 / 26 & {\bf 58.8} & {43.1} & {\bf 62.6} & {\bf 74.6} \\
\bottomrule
    \end{tabular}
}
    \caption{{\bf Object detection performance with DINO on COCO val2017.} AP$^S$, AP$^M$, and AP$^L$ indicate the results for small, middle, and large bounding boxes. FPS metrics are reported based on single-scale testing. FlashInternImage, when integrated with DCNv4, not only achieves superior performance but also maintains a competitive inference speed. All experimental results were obtained with our codebase.} 
    \label{tab:detection_app}
\end{table}

\begin{table*}
    \centering
\resizebox{0.9\linewidth}{!}{
\setlength{\tabcolsep}{9pt}
\begin{tabular}{l|r|r|r|r|r}
    \toprule
    \multicolumn{1}{c|}{\multirow{2}{*}{Operator}} & \multicolumn{5}{c}{Runtime (ms)} \\ 
     & $56 \times 56 \times 128$ & $28 \times 28 \times 256$ & $14 \times 14 \times 512$ & $7\times7\times1024$ & $14 \times 14 \times 768$ \\
    \hline    
      Conv ($3 \times 3$) & 0.833 / 0.596 & 0.708 / 0.464 & 0.653 / 0.384 & 0.687 / 0.459 & 1.31 / 0.754  \\
      Attention (torch)   & 32.0 / ~~20.3 & 4.06 / ~~2.66 & 1.01 / 0.801 & 0.789 / 0.357 & 2.28 / ~~1.44 \\
      FlashAttention-2 & N/A / ~~2.81 & N/A / 0.641 & N/A / 0.256 & N/A / 0.209 & N/A / 0.451 \\
      Window Attn ($7\times 7$) & 4.48 / ~~1.87 & 2.39 / ~~1.00 & 1.35 / 0.581 & 0.824 / 0.371 & 2.12 / 0.911  \\
      DCNv3  & 3.28 / ~~2.98 & 1.62 / ~~1.42 & 0.846 / 0.748 & 0.526 / 0.546 & 1.37 / ~~1.10 \\
      \ourmethod{} (lightweight) &  0.762 / 0.547 & 0.419 / 0.313 & 0.375 / 0.192 & 0.306 / 0.130 & 0.389 / 0.249\\
      \ourmethod{} & 1.28 / 0.873 & 0.738 / 0.483 & 0.452 / 0.324 & 0.334 / 0.265 & 0.787 / 0.463\\
      \bottomrule
    \end{tabular}
}
    \caption{{\bf Module-level operator benchmark on standard input shape with various downsample rates.} 
    }
    \label{tab:op_module_low_res}
    
\end{table*}

\begin{table}
    \centering
\resizebox{\linewidth}{!}{
\setlength{\tabcolsep}{8pt}
    \begin{tabular}{l|c|cc|c}
    \toprule
    	\multirow{2}{*}{Model} & crop & \multirow{2}{*}{\#param} & \multirow{2}{*}{\begin{tabular}{@{}c}
     FPS
    	\end{tabular}} & mIoU \\
    	& size & & & (SS)   \\
    	\hline
      Swin-T & $512^2$ & 47.4M & ~96 / 146& 49.6 \\
      ConvNeXt-T & $512^2$ &47.7M & 111 / 155 & 49.6 \\ %
      InternImage-T &$512^2$ & 48.6M  & ~91 / 123 & 50.6\\
      \rowcolor{gray!20}
      FlashInternImage-T & $512^2$ & 48.5M & 105 / 145 & {\bf 51.3}\\
      \hline 
      Swin-B  &$640^2 $&  110M& 48 / 71 & 51.9\\
      ConvNeXt-B  & $640^2 $& 111M &54 / 77  & 50.7\\ 
      InternImage-B &$640^2 $&  118M &45 / 58 & 52.1 \\
        \rowcolor{gray!20}
        FlashInternImage-S & $640^2 $& 71.5M & 60 / 80 & {52.6}\\
    \rowcolor{gray!20}
      FlashInternImage-B & $640^2 $& 118M & 55 / 75 & {\bf 53.4} \\
      \hline
      Swin-L & $640^2$ &  218M &37 / 55& 56.1 \\
      ConvNeXt-L  &$640^2$ & 220M & 41 / 60 & 55.7 \\
      InternImage-L* & $640^2$ &242M  & 33 / 45 & 55.5 \\
      \rowcolor{gray!20}
      FlashInternImage-L  & $640^2$ &242M  & 41 / 59 & {\bf 56.7} \\
    \bottomrule
    \end{tabular}
}
    \caption{{\bf Semantic segmentation performance on the ADE20K validation set.} All models are trained with Mask2Former. ``SS''  denotes single-scale testing. FPS is reported with single-scale testing. FlashInternImage w/ DCNv4 exhibits a significant advantage in terms of performance while also maintaining competitive inference speed. *: In this experiment, we observed a significant overfitting phenomenon.}
    \label{tab:segmentation_app}
\end{table}

\paragraph{Downstream results for other backbones:}

As demonstrated in \cref{tab:other_backbone_downstream}, we evaluated the performance of ``ConvNext+DCNv4" and ``ViT+DCNv4" on downstream tasks, specifically focusing on semantic segmentation tasks using the UperNet head. Our observations indicate that substituting the previously used DWConv or Attention with our DCNv4 leads to an increase in inference speed. For ConvNext, using DCNv4 rather than DWConv also achieves higher performance.

\begin{figure*}
    \centering
    \includegraphics[width=0.24\linewidth]{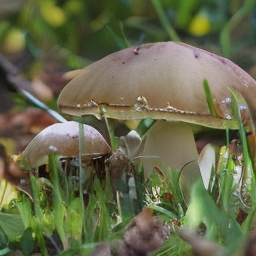}
    \includegraphics[width=0.24\linewidth]{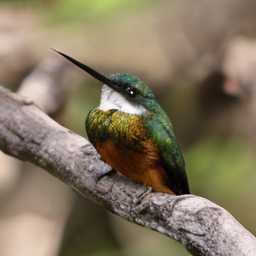}
    \includegraphics[width=0.24\linewidth]{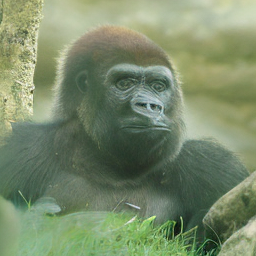}
    \includegraphics[width=0.24\linewidth]{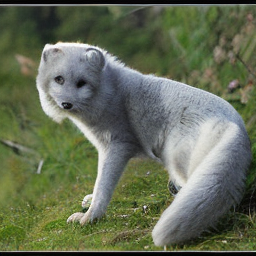}
    \includegraphics[width=0.24\linewidth]{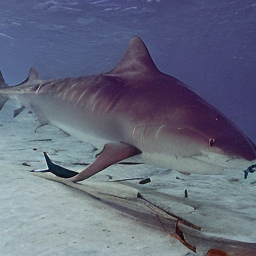}
    \includegraphics[width=0.24\linewidth]{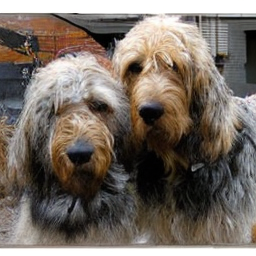}
    \includegraphics[width=0.24\linewidth]{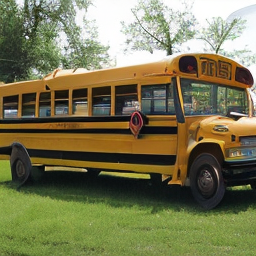}
    \includegraphics[width=0.24\linewidth]{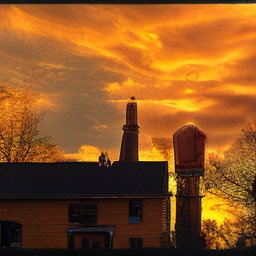}
    \caption{ImageNet $256\times 256$ generation results of U-Net + \ourmethod{} latent diffusion model.}
    \label{fig:diffusion_gen}
\end{figure*}

\paragraph{Visualization for image generation:} We show qualitative results of our latent diffusion model with \ourmethod{} in Fig.~\ref{fig:diffusion_gen} for a better illustration. \ourmethod{} also can work well in this generation task.

\begin{table}
    \centering
    \resizebox{\linewidth}{!}{
    \setlength{\tabcolsep}{20pt}
    \begin{tabular}{l|cc}
    \toprule
    Model & mIoU & FPS \\
    \hline
       ConvNeXt-B & 49.1 & 95 / 147\\
      \rowcolor{gray!20}
       ConvNeXt-B + \ourmethod{}  &49.9& 96 / 164 \\
       \hline
       ViT-B & 48.8& 51 / ~74\\
      \rowcolor{gray!20}
       ViT-B + \ourmethod{} &48.8&67 / ~92\\
        \bottomrule
    \end{tabular}}
    \caption{\textbf{\ourmethod{} in other architecture.} 
    All models are trained with UperNet on ADE20K. By replacing ConvNext's DWConv with DCNv4, we not only achieved better inference speeds but also obtained improved results. Replacing the Attention in ViT with DCNv4 resulted in faster inference speeds while maintaining the same level of performance.}
    \label{tab:dropin_convnext_vit}

    \label{tab:other_backbone_downstream}
\end{table}

\begin{table*}
    \centering
\resizebox{0.9\linewidth}{!}{
    \begin{tabular}{l|r|r|r|r|r}
    \toprule
    \multicolumn{1}{c|}{\multirow{2}{*}{Operator}} & \multicolumn{5}{c}{Runtime (ms)} \\ 
     & $200 \times 320 \times 128$ & $100 \times 160 \times 256$ & $50 \times 80 \times 512$ & $25\times 40\times 1024$ & $64 \times 64 \times 768$ \\
    \hline             
      Conv ($3 \times 3$) &  0.602 / 0.234 & 0.623 / 0.199 & 0.725 / 0.214 & 0.422 / ~~0.281 & 0.532 / 0.318 \\
      Attention (torch)   & OOM / OOM & 26.0 / ~~13.1 & 3.05 / ~~1.99 & 0.599 / ~~0.433 & 4.47 / ~~2.69\\
      FlashAttention-2    & N/A / ~~13.4 & N/A / ~~1.81 & N/A / 0.354 & N/A / ~~0.158 & N/A / 0.531\\
      Window Attn ($7\times 7$)  & 1.49 / 0.657 & 0.867 / 0.377 & 0.539 / 0.274 & 0.409 / ~~0.276 & 0.920 / 0.410\\
      DCNv3   &  1.06 / 0.964 & 0.576 / 0.560 & 0.451 / 0.459 & 0.387 / ~~0.409 & 0.534 / 0.543\\
      \ourmethod{}  (lightweight)  & 0.283 / 0.207 & 0.187 / 0.143 & 0.134 / 0.110 & 0.105 / 0.0912 & 0.180 / 0.125 \\
      \ourmethod{}   & 0.446 / 0.346 & 0.313 / 0.272 & 0.267 / 0.270 & 0.268 / ~~0.253 & 0.325 / 0.271 \\
      \bottomrule
    \end{tabular}
}
    \caption{{\bf Module-level operator benchmark on high-resolution input shape with various downsample rates.} 
    }
    \label{tab:op_module_high_res}
\end{table*}

\paragraph{Module-level speed benchmark:}
We show module-level speed benchmark results in Tab.~\ref{tab:op_module_low_res} and~\ref{tab:op_module_high_res}, where additional linear projection layers in each operator are also considered. 
We also include the regular convolution with a $3\times 3$ window here as it mixes channel information.
For \ourmethod{}, we show two variants: 
where the first implementation removes the input/output projection layers inside the module (denoted as lightweight). 
We use this implementation in the ``ConvNeXt + DCNv4'' experiments as it shares similar properties (only performs spatial aggregation) and the amount of computation/parameters as the original depthwise convolution in ConvNeXt.
The second implementation includes the input/output linear projections and is used in the rest of the models described in the paper.

{
    \small
    \bibliographystyle{ieeenat_fullname}
    \bibliography{main}
}

\end{document}